\documentclass[11pt]{article}

\usepackage[final]{acl}
\usepackage{times}
\usepackage{latexsym}
\usepackage{algorithmic}
\usepackage{graphicx}
\usepackage{textcomp}
\usepackage{xcolor}
\usepackage{booktabs}
\usepackage{subcaption}
\usepackage{changepage}

\usepackage{booktabs}
\usepackage{multirow}
\usepackage{caption}

\usepackage[compact]{titlesec}
\titlespacing{\section}{0pt}{1ex}{0.5ex}
\titlespacing{\subsection}{0pt}{1ex}{0.5ex}
\titlespacing{\subsubsection}{0pt}{0.5ex}{0ex}

\usepackage[T1]{fontenc}

\usepackage[utf8]{inputenc}

\usepackage{microtype}

\usepackage{inconsolata}

\usepackage{graphicx}

\title{SAUCE: Summary Analysis Using Conversation Entailment}

\author{Man-Ling Sung, Hemanth Kandula, Jeff Ma, William Hartmann, Matthew Snover
 \\
 RTX BBN Technologies \\ Cambridge MA, USA
 }

\begin{document}
\maketitle
\begin{abstract}
With the growing need for evaluating Large Language Models (LLMs) and their applications to speech, challenges persist in summarizing and evaluating conversations that lack a clear end goal. We introduce SAUCE -- a reference-free, fact-based evaluation pipeline for cross-lingual conversational speech summarization. It measures the accuracy and the fact coverage of a summary through the entailment between conversation and text. 
We compare SAUCE against several popular summarization metrics and demonstrate the effectiveness of capturing information loss due to transcription and translation error and identifying broken summaries. Crucially, unlike black-box LLM evaluators or dense embedding metrics, SAUCE is inherently explainable: it maps summary scores to discrete, verifiable facts, allowing users to pinpoint exact hallucinations or omissions. We illustrate how this interpretability helps developers systematically profile LLM behaviors and gives end-users an actionable tool to verify summary accuracy in noisy, real-world conditions. Preliminary investigations show SAUCE strongly align with human judgment.

\end{abstract}

\section{Introduction}



Machine-generated summarization quality has rapidly improved with the advent of large language models (LLMs) and is commonly used by users for tasks like understanding research papers and generating meeting notes. In some cases, LLM-based summaries have been found to be more accurate than reference summaries in existing academic datasets \cite{zhang2024benchmarking}. Evaluation of summary quality remains a critical challenge, as there is no single true summary for a given document; instead, a distribution of possible summaries can be considered correct \cite{jung2024augsumm}. Users also have their own preferences in terms of both style and content. Any variation in user preference would change the corresponding reference summary, making it impossible to generate sufficient reference summaries to encapsulate these variations. Most existing metrics and measurement pipelines lack the ability to provide quantitative analysis, making it difficult for end users to perform a systematic analysis of summary quality. 
In the scenario of daily conversation---where the end goal and information‑flow structure are often unclear---it becomes even harder to evaluate summary quality.
A quantitative measurement pipeline for such data will assist end users to efficiently evaluate summaries and provide developers with constructive direction for model selection and improvement.


We believe that it is necessary to create quantitative reference-free evaluation metrics in order to accommodate the flexibility of the task. As with prior work, we agree that a single metric cannot capture all of the necessary dimensions \cite{gehrmann2023repairing}. When the scope and end goal are not clearly defined, we believe it is important for the summary to include as much key information as possible and be factually accurate, given the word limit. Therefore, we propose an evaluation pipeline---Summary Analysis Using Conversation Entailment (SAUCE), that captures two specific aspects of the summary: Faithfulness and Coverage, where 
faithfulness captures whether the information in the summary is supported by the text (factual accuracy) and 
coverage captures whether the important information in the text is captured by the summary (coverage of key information). 

We consider the specific use case of evaluating cross-lingual conversational speech summarization. This is a challenging task where human-generated reference summaries do not exist. It's also under-researched compared to traditional text problems. The content is less structured and goal-oriented, unlike meeting recordings or voice messages. Furthermore, conversational speech presents unique challenges beyond simply lacking a clear end goal. Spontaneous dialogues are riddled with disfluencies, speaker overlaps, and topic shifts. When these complexities are compounded by cascading errors from Automatic Speech Recognition (ASR) and Machine Translation (MT), the semantic surface of the text is fundamentally altered. In these noisy conditions, a critical limitation of both traditional n-gram metrics and modern LLM-as-a-judge frameworks is their lack of explainability. When a summary receives a low score, end-users are left guessing whether the penalty stems from a hallucinated entity, a missed core topic, or a stylistic bias. A quantitative measurement pipeline must not only score a summary but also provide a transparent audit trail of why the summary succeeded or failed. SAUCE addresses this by providing a transparent, fact-by-fact ledger. Every 
score is entirely traceable to atomic facts, making the evaluation both interpretable and actionable.

Our contributions are to provide a summarization evaluation pipeline that can: 
1) effectively evaluate cross-lingual conversations in the presence of ASR and MT errors, 2) score factually without human annotation and 3) provide interpretability and quality measurement when the end goal of the text is not clearly defined.

\section{Related work} \label{sec:related_work}

The two major directions for summarization evaluation are \emph{reference-based} and \emph{reference-free}. Reference-based metrics are the more traditional approach and require one or more reference summaries---typically generated by humans---for comparison. ROUGE \cite{lin2004rouge} and BERTScore \cite{zhang2019bertscore} are popular examples. They compute a similarity score for matching word units or tokens, sentence by sentence between the candidate summary and reference summary. While widely used in the community, research shows they only measure topic presence rather than factual accuracy in the summary \cite{deutsch2021understanding}. ROUGE is also not sensitive enough to the impact of major errors in the input, like those from ASR or MT errors \cite{nelson2024cross}.

Due to both the limitations and the difficulties of reference-based approaches, a number of reference-free metrics have been proposed as well. One general line of work is on measuring factual consistency through text entailment. Text entailment, as known as Natural Language Inference (NLI), measures how the hypothesis and generated text align or contradict each other. SummaC \cite{laban2022SummaC}, a popular example, measures the coherence between the document and summary by segmenting the document into sentence units and scoring document and summary sentence pairs with text entailment models. Its better performing variant SummaC\textsubscript{Conv} is heavily trained on a specific text domain, which may not generalize to our cross-lingual conversation setting. FENICE extends the evaluation capability to long-form text summarization with claim extraction and entailment, follow by coreference resolution \cite{scire2024fenice}.


Another approach to reference-free evaluation is to directly use LLMs for scoring \cite{akkasi2023reference}, such as GPTScore \cite{fu2024gptscore}. This "LLM-as-a-judge" paradigm typically involves prompting high-capacity models like GPT-4 to evaluate summaries across various dimensions of interest. G-Eval \cite{liu2023g} pioneered this by using chain-of-thought (CoT) prompting to align model scores with human judgment, while Prometheus \cite{kim2023prometheus} and JudgeLM \cite{zhu2023judgelm} introduced fine-tuned, open-source alternatives designed for reproducibility. FineSurE \cite{song2024finesure} extends the work to include LLM to extract key facts for LLM scoring. 
While these frameworks can provide qualitative explanations, they are difficult to interpret quantitatively and become computationally infeasible when scaling to the massive/gigantic datasets typical of real-world applications. Furthermore, these evaluators are susceptible to significant systematic biases, self-preference bias \cite{panickssery2024llm}, verbosity bias, where longer summaries are unfairly favored \cite{hu2024llm}, and a recently identified overlap bias where LLMs increasingly favor AI-generated summaries over human-written ones \cite{fang2026blind}. Most critically, LLM judges often overlook granular factual contradictions in favor of surface-level fluency \cite{chen2024humans, fu2024gptscore}, which further motivates our need for the explicit, fact-extraction and entailment pipeline proposed in this work. Fact checking of the extracted facts against an external knowledge database is not in the scope of this work.

\section{Proposed approach} 
\label{sec:proposed_approach}

SAUCE evaluates faithfulness and coverage of the summary with 2 metrics: SAUCE\textsubscript{Faithful} and SAUCE\textsubscript{Coverage}. SAUCE\textsubscript{Faithful} measures the percentage of summary facts entailed by the conversation and SAUCE\textsubscript{Coverage} measures the percentage of conversation facts covered in the summary. It can be seen as an agentic workflow that uses an LLM to extract key information from the text. 

\subsection{Pipeline Design} \label{sec:general_metric}

\begin{figure}[th]
  \centering
  \includegraphics[width=1\linewidth]{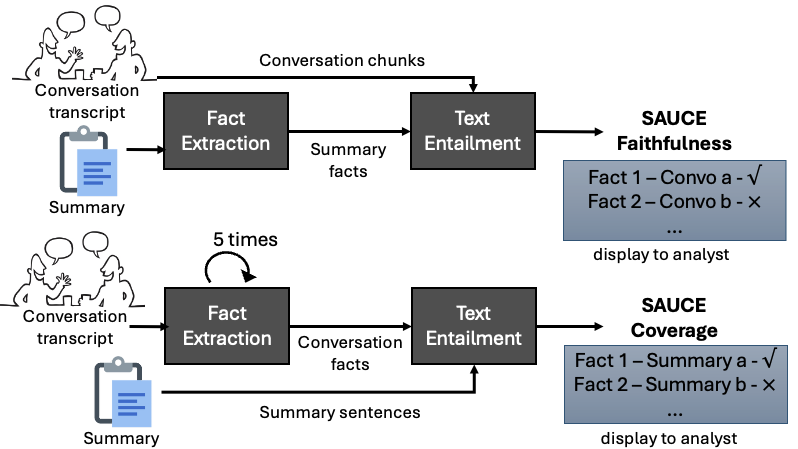}
  \caption{Workflow of the SAUCE. The user can access the fact-text pairs to understand the missing or wrong information in the summary with corresponding faithfulness and coverage scores.}
  \label{fig:metric_design}
\end{figure}

SAUCE consists of fact extraction followed by text entailment. Illustrated in Figure \ref{fig:metric_design}, we first extract atomic facts from the conversation/summary with GPT-4o (gpt-4o-2024-08-06), then determine if the facts are entailed by the summary/conversation using DeBERTa-v3 (deberta-v3-large-mnli). While LLMs are prone to hallucination, they have been shown to outperform other models with about 90\% accuracy in the task of fact extraction \cite{polak2024extracting, dagdelen2024structured}. Our internal evaluation found the extracted facts to be over 98\% accurate. We use DeBERTa-v3 as the text entailment model because of its state‑of‑the‑art performance in this task \cite{he2021debertav3} and its limited computational requirements. 

To calculate SAUCE\textsubscript{Faithful}, after extracting all the facts in the summary, the facts are scored against chunks of the conversation, with 5 lines of conversation per chunk. If a fact-to-chunk pair scores higher than a threshold, then the fact is considered to be entailed by the conversation. The score is calculated as the proportion of matched facts. We also experimented with entailing summary facts against facts extracted from the conversation, but found that the proposed fact‑to‑chunk approach yields significantly better performance. 

For SAUCE\textsubscript{Coverage}, we ask the LLM to only extract the N most important facts from the conversation, where N is defined by the length and desired content compression rate of the task. We do not ask for all the facts in the conversations because the LLM will frequently generate over 100 facts from a 10-minute recording, defeating the purpose of summarizing the conversation. Importance is subjective, but in the case of a topic-defined conversation, the LLM can be guided towards preferring certain topics and information through prompting. To overcome the nondeterministic output from the LLM, we extract the N most important facts 5 separate times and concatenate the list for scoring. 
We also assume that facts that are selected multiple times are more likely to be important. For text entailment, each fact is scored against each line of the summary. This score is calculated as the proportion of matched facts. 

A key advantage of SAUCE is its transparent by-product: the fact ledger. Rather than outputting a single opaque float value, the system surfaces the exact list of atomic facts that were successfully entailed and those that were contradicted or unsupported. For end-users, this ledger acts as a rapid verification tool to check a summary against the source without re-reading the entire transcript. For developers, it isolates errors of hallucinations, captured by unentailed summary facts, from errors of omission (missing key topics, captured by unentailed conversation facts).

\subsection{Performance on Conversation Entailment} \label{sec:convo_entail}

Fact-to-fact (sentence-to-sentence) entailment has been widely studied in the text community. However, much of the work has focused on written text as opposed to conversational speech. In order to verify whether DeBERTa-v3 is suitable for fact-to-conversation entailment, we evaluate the setup on a publicly available conversation entailment dataset from MSU \cite{zhang2009we}. It consists of hypothesis statements and supporting conversation chunks from Switchboard\footnote{https://catalog.ldc.upenn.edu/LDC97S62} extracted by human annotators. Conversation entailment is a challenging problem; only half of all the statements MSU collected are agreed upon by all annotators. Based on our internal analysis, we found many instances where the entailment was questionable, or the statements were only entailed when considering the conversation as a whole. 

We set up two experiments to evaluate our conversation entailment setup. First, we directly entail the conversation chunks and statements in the dataset with DeBERTa-v3. Second, we take the whole conversation that each chunk belongs to and run our SAUCE\textsubscript{Faithful} scoring on it; the most entailed conversation chunks are scored. We experimented with a threshold of 0.9, 0.95 and 0.98. Results are shown in Table \ref{tab:convo_entail}.

\begin{table}[]
  \centering
\begin{tabular}{p{0.10\linewidth} cccccc}
\toprule
             & \multicolumn{3}{c}{\textbf{Selected chunk}} & \multicolumn{3}{c}{\textbf{Full conversation}} \\
            \midrule
Thres    & 0.9             & 0.95           & 0.98           & 0.9         & 0.95       & 0.98       \\
            \midrule
All      & 77.0 & 79.2 & 77.0 & 68.1 & 71.4 & 74.2 \\
NoEnt. & 63.6 & 71.6 & 77.1 & 39.7 & 49.9 & 61.4 \\
Ent.     & 88.4 & 85.7 & 80.0 & 92.2 & 89.7 & 85.0 \\  
    \bottomrule
\end{tabular}
\caption{Accuracy rate (\%) on conversation entailment. \emph{Ent.} - entailed pairs, \emph{NoEnt.} - non-entailed pairs.}
\label{tab:convo_entail}
\end{table}

\begin{table*}[ht]
\begin{tabular}{p{0.30\linewidth}  p{0.65\linewidth}}
\toprule
Extracted Fact                                  & Chunked conversation         \\
\midrule
B's mother got in trouble for getting sick & B: Oh no, I can remember my mother getting in trouble if, you know, one of was sick, and I know she probably didn't make hardly anything, you know, compared to the work that she did.       \\
A plays on MURPHY BROWN                    & A: He, he plays {[}laughter{]} on   MURPHY BROWN.   \\           \bottomrule 
\end{tabular}
\caption{Examples of non-entailed pairs that are mis-classified as entailed. The first row confuses factual and hypothetical scenarios, which is also confusing to humans. Second row confuses the subject, the speaker vs. the person indicated by `he'.}
\label{tab:non-entail-exp}
\end{table*}

Non-entailed pairs have worse performance than entailed pairs. Most of the mis-entailed statements are those where the tenses or identities of speakers are swapped (see Table \ref{tab:non-entail-exp}). This reflects that even with acceptable performance, text entailment models are not tailored made for conversations entailment. We also noticed the dataset has annotation issues; some of the selected conversation chunks from the entailed pairs do not support the statement in the data, but other parts of the conversation do support the statement. For the remainder of our study, we select a threshold of 0.9 as we believe it is more critical to correctly detect entailed pair when it is present. 

\section{Evaluation of metrics on summarized ASR+MT conversations} \label{sec:experiment_asr_mt}

To evaluate our proposed metrics, we analyze performance on a number of scenarios. The first evaluation is to compare the performance on the conversational speech summarization task with transcription and translation errors.

\subsection{Datasets} \label{sec:asr_dataset}

We experiment on conversational telephone speech (CTS) across 4 languages, Arabic, Spanish, Mandarin and Russian. Audio and transcripts are available through the LDC (Linguistic Data Consortium) for all languages. CallHome\footnote{https://catalog.ldc.upenn.edu/\{LDC97T19, LDC97S45\}} is used for Arabic, CallHome\footnote{ 
 https://catalog.ldc.upenn.edu/\{LDC96T17, LDC96S35\}} and Fisher\footnote{https://github.com/joshua-decoder/fisher-callhome-corpus} for Spanish, HUB5\footnote{https://catalog.ldc.upenn.edu/LDC98S69} for Mandarin and
Mixer 3 Speech\footnote{https://catalog.ldc.upenn.edu/LDC2023S02} for Russian. Crowd-sourced English translations are publicly available for Arabic \cite{kumar2014translations} and Spanish \cite{post2013improved}.
Mandarin \cite{wotherspoon2024advancing} and Russian are our in-house translations. 
Each conversation is 10 minutes long in average. The selection of CTS data fits the domain of loosely defined and structured conversation, which SAUCE is designed to handle. CallHome and CallFriend corpora contain conversations with family and friends. There are no clearly defined goals and speakers can discuss anything. The Fisher corpus defines topics to discuss, but there is no requirement for the speakers to come to any conclusion.




\subsection{Experiment setup}

\textbf{Summary:} We prompt GPT-4o to generate a 200-word summary for each conversation.
We specifically prompt the model to generate concise and comprehensive summaries with clarity, based only on the provided conversation and to include the main ideas and essential information. 
We experimented with a range of prompts, but they have a limited impact on the overall quality of the summaries when there is no topic specified.
We tested various word limits ranging from 50, 100, 200 and 500 words, and found 200 words to be a good balance considering the length of the conversation and readability to a human user. The LLM is forced to downselect information included in the summary.

\textbf{ASR+MT pipeline:} We use a cascaded ASR+MT pipeline to obtain the English conversations for summarization containing ASR+MT errors. We use Whisper-large-v3 \cite{radford2023robust} for ASR and NLLB-200-distilled-1.3B \cite{costa2022no} for MT. Both models are fine-tuned with the training set of each specific language. While we use a cascaded approach for the purpose of measuring the impact of ASR and MT errors, our metrics would be equally applicable to direct speech summarization approaches \cite{shang2024end}. 

\textbf{Evaluation:} All the dev and test sets of each language are used for evaluation. We set N=20 for conversation fact extraction for evaluating 200 words summary. We compared our metrics along with SummaC and FENICE, NLI-based metrics that is similar with SAUCE, as well as LLM-as-judge metrics: G-EVAL and FineSurE.

For our metrics, each summary is scored against their ground truth conversation chunks for SAUCE\textsubscript{Faithful} and facts extracted from the ground truth conversation for SAUCE\textsubscript{Coverage}. 
All the other metrics score the summary according to how their pipelines are setup, either against the ground truth conversation (SummaC, G-EVAL) or its extracted facts (FENICE, FineSurE). 
For SummaC, we used the recommended SummaC\textsubscript{Conv} (Convolution of the NLI sentence pair matrix) and the traditional SummaC\textsubscript{ZS} (Zero-Short model). To ensure a fair evaluation on conversation entailment, we expanded the granularity to 5 sentences to match our SAUCE\textsubscript{Faithful} scoring setup.



\subsection{Results on summary evaluation}

We present the average of the ASR and MT scores in WER (Word Error Rate, the lower the better) and BLEU along with the summary scores on the ground truth, MT and ASR+MT translations in Figure \ref{fig:asr_mt_summ_score}. Breakdown of the scores are in Appendix \ref{sec:asr_mt_detail}.

\subsubsection{Comparison with NLI-based metrics}

\begin{figure*} [ht]
    \vskip -1em
    \centering
    \includegraphics[width=0.9\linewidth]{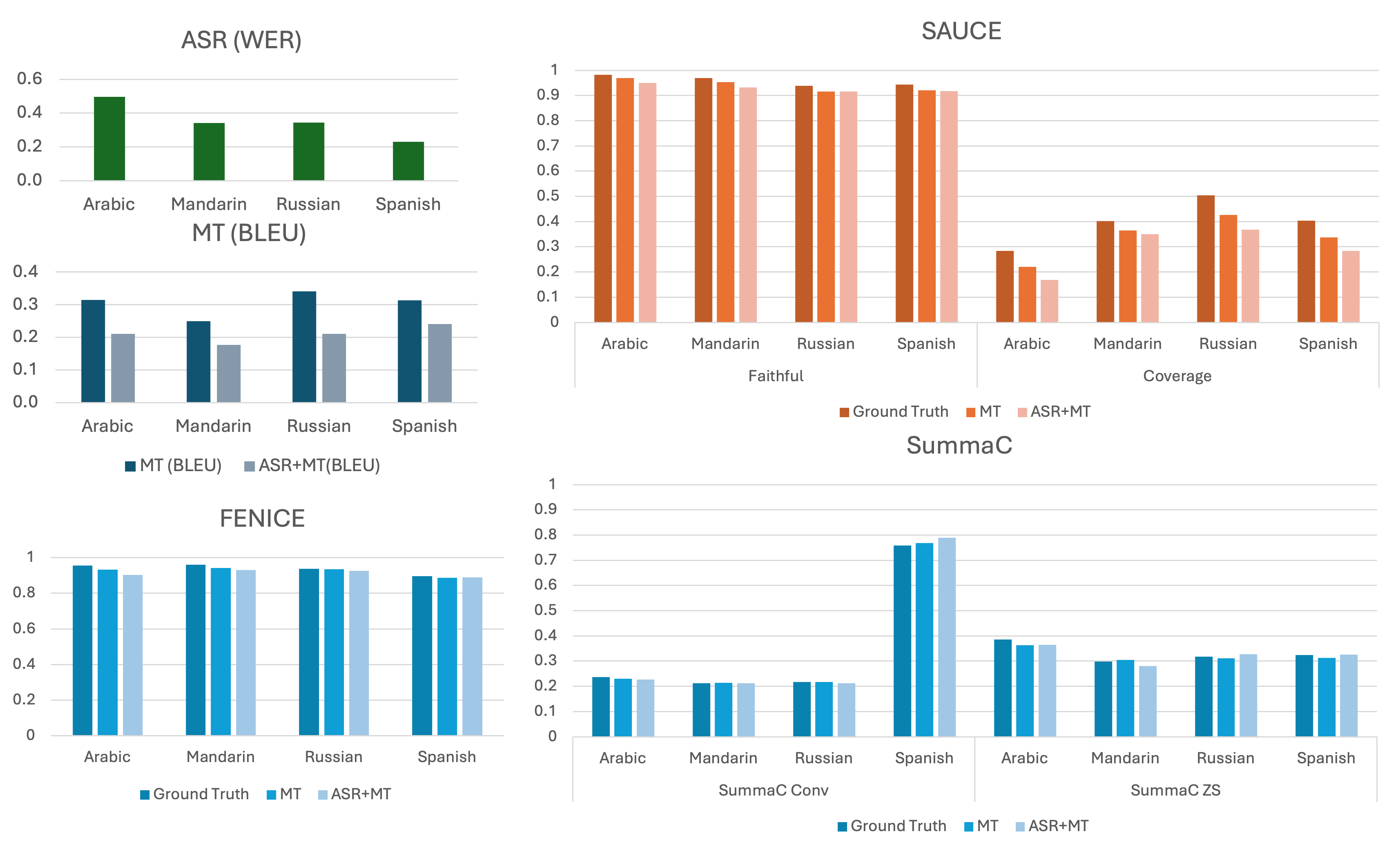}
        \vskip -1em
    \includegraphics[width=1\linewidth]{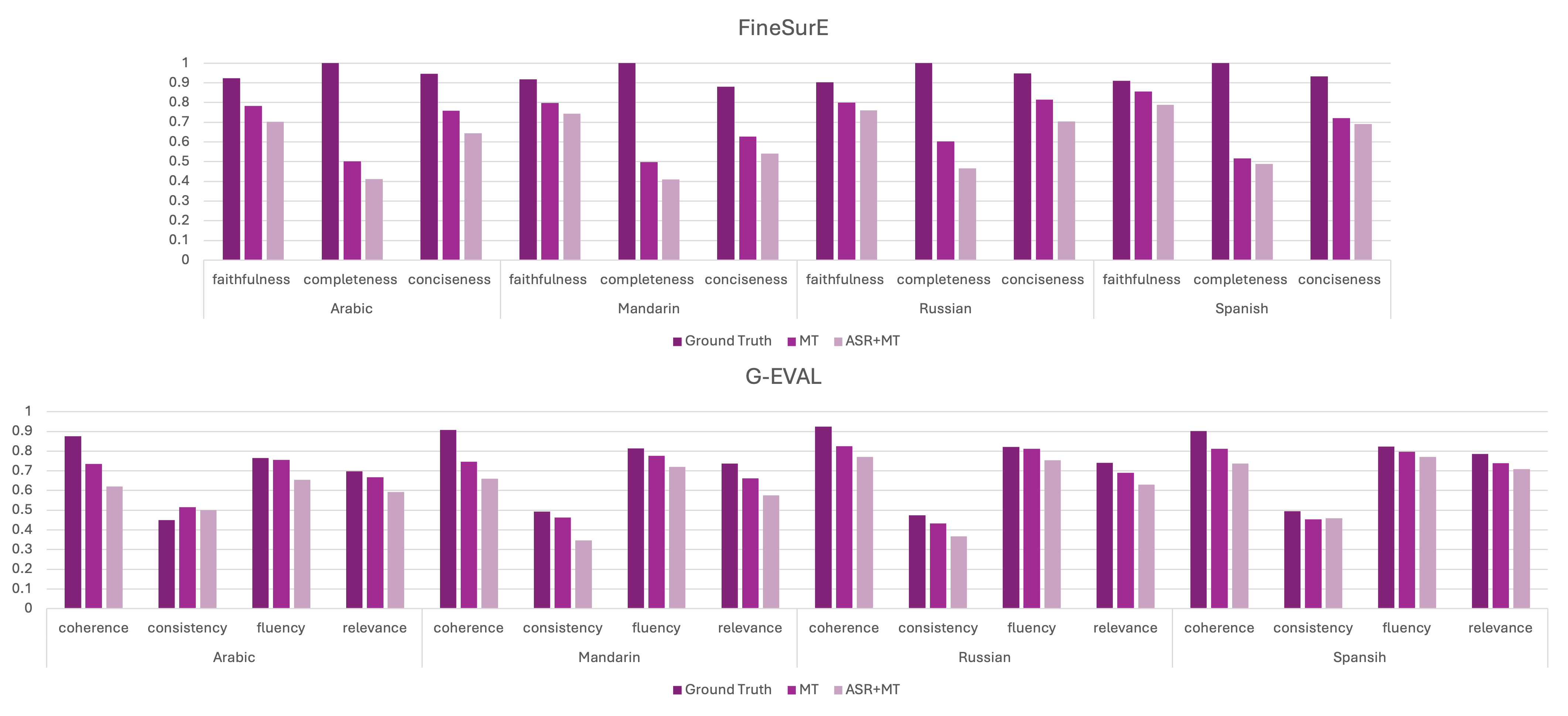}
    \caption{Mean WER and BLEU scores of each language, and their summary scores on ground truth, MT and ASR+MT English transcriptions. We present the plots in orange for SAUCE, blue for NLI-based metrics, purple for LLM-as-judge metrics.}
    \label{fig:asr_mt_summ_score}
\end{figure*}

   
Among the NLI-based metrics, SAUCE is the only metric that have a consistent drop when MT and ASR+MT errors are introduced, while the SummaC metrics do not have a consistent pattern on different languages, and FENICE has very limited degradation especially on Spanish.

Given the high WER and its contribution to a large drop in BLEU (8-12\% absolute), we expect a large impact on summarization quality. SAUCE\textsubscript{Coverage} is nearly cut in half when both ASR and MT errors are introduced---reflecting the fact that GPT-4o has difficulty extracting information with the degradation in transcription accuracy and translation quality---while there is no significant impact on SummaC\textsubscript{Conv,ZS} and FENICE.
SAUCE\textsubscript{Faithful} and FENICE are more consistent and maintain a score of 90+\%, regardless of the quality of the input. Based on our investigation, we find that GPT-4o generates more generic summaries in the presence of these errors. The lost of information is captured by SAUCE\textsubscript{Coverage}. However, FENICE lacks a "Coverage" metric, making it blind to omitted information, only very limited degradation is observed with the errors introduced. 

SummaC\textsubscript{Conv} sees a 40-60\% absolute difference between Spanish and the other three languages. We suspect the large shift is due to differences in linguistic structure and word usage compare. SummaC is primary designed for written text and might not perform well on conversational style data, especially in languages that differ significantly from the training data.

Both SummaC and FENICE adopt a stricter NLI-based pipeline to avoid LLM errors, but their traditional NLP models and mechanisms, which were built on structured text, completely break down when confronted with the disfluencies, speaker overlaps, and ASR/MT errors typical of natural speech.

\subsubsection{Comparison with LLM-as-judge metrics}

Compared with G-EVAL and FineSurE, consistent degradation is observed in all metrics under MT and ASR errors, except for G-EVAL's consistency in Arabic and Spanish. 
LLM-as-a-judge frameworks like G-EVAL and FineSurE act as computationally expensive "black boxes" that are highly susceptible to biases and tend to favor surface-level fluency over strict factual accuracy, making it difficult to trace exactly why a score was given. Even though FineSurE attempts to ground its evaluation by extracting key facts, it still ultimately depends on the LLM's inherent reasoning to output a score.  Additionally, FineSurE specifically suffers from an empirical flaw in its alignment logic (As shown in figure \ref{fig:asr_mt_summ_score}): when evaluating a Ground Truth (GT) summary against key facts extracted from that exact same GT summary, the LLM fails to score a perfect 100\% in faithfulness, proving that using an LLM to judge semantic alignment is inherently unreliable. The 100\% completeness across all languages, which implies that all key information is included in the GT summaries, seems unrealistic. This is because information is inevitably omitted during summarization, especially in long conversations (e.g., those lasting 30 minutes in the Mandarin dataset). SAUCE resolves these issues by using LLMs only for robust fact extraction and a dedicated NLI model for strict entailment, yielding a transparent, fact-by-fact ledger that reliably measures both faithfulness and coverage despite heavy noise. For the complete dataset and granular sub-metric scoring, see Table \ref{tab:raw_results_full} in the Appendix.

\subsection{Analysis on summaries}
 
We selected conversation sp\_1578 from Spanish CallHome and listed the summaries and SAUCE and SummaC scores in Table \ref{tab:spanish_convo_example}. There is a constant drop in SAUCE\textsubscript{Coverage} and some degradation in SAUCE\textsubscript{Faithful} when errors are introduced. The SummaC\textsubscript{Conv, ZS} metrics do not show the degradation consistently.
We highlight a few sentences, denoted with letters for comparison. The quality of the summary degrades across versions; the overall description of topics in (a) becomes more generic, which is still accurate (match on SAUCE\textsubscript{Faithful}) but loses some information (drop in SAUCE\textsubscript{Coverage}). Facts are also distorted due to transcription and translation errors, causing drop in both SAUCE\textsubscript{Faithful} and SAUCE\textsubscript{Coverage}. For example, (b) Carlos did not receive assistance is distorted to not receiving attention in the MT version to himself showing a lack of interest in ASR+MT. (c) Ernesto, from ``buying a car and learning to drive" to ``planning to purchase a car" with ASR+MT errors. The connection between facts also disappears due to ASR+MT errors, as in (d), which does not hurt SAUCE\textsubscript{Faithful} but hurts SAUCE\textsubscript{Coverage} when the facts that look for correlations between incidents are missing. For example of conversation facts used for scoring, see Appendix \ref{sec:fact_examples}. The metrics make the impact of these cascading errors highly interpretable. When faced with ASR+MT degradation, we observe that the LLM often adopts a conservative generation strategy. It retreats to producing generic, high-level statements that remain factually accurate (maintaining high faithfulness) but strips away the granular, nuanced details of the conversation (causing a sharp drop in coverage). Traditional metrics like SummaC obfuscate this behavioral shift, whereas our fact-level mapping explicitly diagnoses it.
\begin{table*}[]
\small
\centering

\begin{tabular}{p{0.95\linewidth} } 
\toprule
\textbf{Summary from human translation} (SAUCE\textsubscript{Faithful}: 0.94, SAUCE\textsubscript{Coverage}: 0.6, SummaC\textsubscript{Conv}: 0.6,  SummaC\textsubscript{zs}: 0.47)\\ 
\midrule
 In the conversation, two individuals discuss various {\color{blue}(a) personal matters involving health, family, and home renovations}. One of them recently visited Joana to deliver some clothes, and discovered that Isa was sick with a cold. {\color{purple}(b) Carlos, who drove the speaker, received no assistance during the visit.} The conversation reveals that there isn't much improvement in Isa's condition, but there is a hope for becoming more independent. {\color{orange}(c) Ernesto, a family member, has bought a car and is learning to drive}, bringing some cheerful news. The family is also involved in home improvements, including adding a store and planting grass, enhancing their surroundings. However, the speakers acknowledge some ongoing challenges, emphasizing the importance of gratitude and support, especially for Carlos, who is not in a good place mentally. {\color{olive}(d) Carlos’ surgery was postponed due to his cold and allergies, reminiscent of one speaker’s past experience with an adverse reaction to antibiotics, leading to a discussion about the risks of self-medication.}... 
 \\ 
\midrule
\textbf{Summary from MT} (SAUCE\textsubscript{Faithful}: 0.82, SAUCE\textsubscript{Coverage}: 0.44, SummaC\textsubscript{Conv}: 0.74,  SummaC\textsubscript{zs}: 0.43)\\ 
\midrule
In a conversation between two individuals, they discuss various topics, including {(\color{blue}a) personal encounters, family news, and health concerns.} One participant mentions visiting Joana to drop off clothes and finding Isa unwell due to a cold. {\color{purple}(b) There's a discussion about Carlos, who seems to be undergoing challenges but not receiving much attention.} The conversation shifts to family updates, {\color{orange}(c) mentioning Ernesto's new car purchase and driving lessons}, as well as home renovations being done by several friends and family members like Cristi and Gaby. The tone reflects a mixture of envy and gratitude, emphasizing the importance of being thankful despite current hardships. They highlight the significance of family unity, particularly regarding Carlos's need for support amidst his health issues. {\color{olive}(d) Carlos is recovering but has had an operation postponed to November due to sinusitis-like symptoms.} ... 
\\ 
\midrule

\textbf{Summary from ASR + MT} (SAUCE\textsubscript{Faithful}: 0.84, SAUCE\textsubscript{Coverage}: 0.2, SummaC\textsubscript{Conv}: 0.54,  SummaC\textsubscript{zs}: 0.42)\\ 
\midrule
In the conversation, the speakers discuss {\color{blue}(a) various personal and familial updates.} Speaker A mentions visiting Joana to drop off clothes and encountering Isa, who is sick with a cold. Isa's condition and the {\color{purple}(b) lack of interest shown by Carlos, who accompanied A, are highlighted.} The speakers discuss the need for A to reconsider certain decisions, but A resolutely rejects this notion, emphasizing independence and familial unity as priorities. The conversation shifts to news about {\color{orange}(c) Ernesto planning to purchase a car}, while renovations are underway on friends' and family's houses. B praises these efforts but expresses frustration with perceived stagnation in their situation. The speakers highlight gratitude towards God for their inner wellbeing and unity, despite external challenges. A underscores the importance of supporting {\color{olive}(d) Carlos, particularly due to his health issues, which include a postponed surgery due to allergies.} ... 
\\ \hline

\end{tabular} 
\caption{Summaries and scores on Spanish conversation (sp\_1578) on ground truth, and with MT, ASR+MT error. Sentences related to the same set of facts are colored and labeled with corresponding alphabets.} 
\label{tab:spanish_convo_example}
\end{table*}

\subsection{Experiment on metrics correlation}

We also evaluate the correlation of the conversation quality and metrics at the document level. We believe that with ASR and MT errors introduced, the quality of the transcription should degrade so as the summary quality. Therefore, we measured the Kendall pair-wise rank correlation coefficient \cite{abdi2007kendall} of ground truth and ASR+MT summary pairs against the metrics, assuming more error corresponds to worse summary quality, thus lower scores. In Table \ref{tab:tau_asr_mt}, SAUCE\textsubscript{Faithful} and SAUCE\textsubscript{Coverage} are highly correlated with conversation quality, with SAUCE\textsubscript{Coverage} being more correlated than SAUCE\textsubscript{Faithful}. While SummaC\textsubscript{Conv,ZS} shows little to no correlation.

\begin{table*}[t] \centering
\begin{tabular}{c|cccc} \hline
\toprule
Metric         & SAUCE\textsubscript{Faithful} & SAUCE\textsubscript{Coverage} & SummaC\textsubscript{conv} & SummaC\textsubscript{zs} \\ 
\midrule
Kendall rank coef. & 0.153    & 0.677 & -0.266       & 0.008      \\    
\bottomrule
\end{tabular} 
\caption{Tau coefficient of metrics on ground truth and ASR+MT conversations. Assumes summaries generated from ground truth are better than summaries generated from ASR+MT output.} \label{tab:tau_asr_mt}
\end{table*}

\section{Broken summary evaluation} \label{sec:broken_summary}


A robust system should be able to detect improperly formatted or unreadable (garbage) summaries. We experimentally test whether evaluations metrics can identify them.

\subsection{Experiment setup}

\begin{figure} [ht]
    \centering
    \vskip -0.5em
       \includegraphics[width=\linewidth]{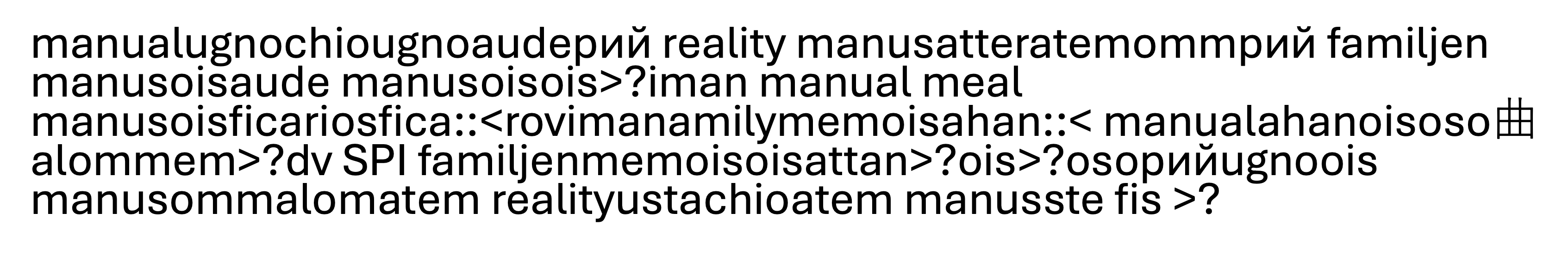}
    \caption{Example of broken summary.} \label{fig:broken_summary}
    \vskip -1em
\end{figure}

CallHome Spanish and Arabic Eval sets are used for this experiment. Normal 200-word summaries are generated by off-the-self Mistral7B-instruct-v0.3 \cite{jiang2023mistral} on the ground truth conversations. Broken summaries (example in Figure \ref{fig:broken_summary}) are generated by fitting the same conversations and prompt to an over-compressed Mistral7B model with td2 quantization. The summaries are scored against the ground truth conversation with SummaC\textsubscript{Conv, ZS}, and SAUCE\textsubscript{Faithful, Coverage}.

\subsection{Results}
 
In Table \ref{tab:broken_summary}, when applying different scoring metrics on the broken Spanish summaries, the scores in SummaC\textsubscript{Conv} are still higher than normal Arabic summaries. In our setup, it would be possible to identify such model dysfunction with the exceptionally low SAUCE\textsubscript{Faithful} and SAUCE\textsubscript{Coverage}.

\begin{table*}[th]

\centering
\begin{tabular}{cccccc}
\toprule
Metrics       &  SummaC\textsubscript{Conv}  & SummaC\textsubscript{ZS} & SAUCE\textsubscript{Faithful} & SAUCE\textsubscript{Coverage} \\ \midrule
Spanish\textsubscript{norm} &   0.641       & 0.307     & 0.941     & 0.358  \\
Spanish\textsubscript{broke} &  0.598       & 0.104     & 0.356     & 0.000  \\
Arabic\textsubscript{norm}  &  0.221       & 0.395     & 0.980     & 0.363 \\
\bottomrule
\end{tabular}
\vskip -0.5em
\caption{Metrics on Eval sets. Spanish\textsubscript{norm} and Arabic\textsubscript{norm} are normal summaries, Spanish\textsubscript{broke} is broken summaries.}
\label{tab:broken_summary}
\end{table*}

\section{Evaluation of off-the-shelf LLMs} \label{sec:off_the_shelf_LLMs}

Lastly, we would like to know, given a set of models to choose from, how SAUCE can help end users to understand the models' behaviors and identify a suitable model for their tasks.
\subsection{Experiment Setup}

We compare various existing LLMs using our proposed metrics, namely GPT-4o, Mistral7B-instruct-v0.3, DeepSeek-R1 \cite{guo2025deepseek} and its distilled version using the Qwen2.5 model \cite{hui2024qwen2} (DeepSeek-R1-Distill-Qwen-14B). We experimented with various prompts on different models, but maintained the same prompt in Section \ref{sec:experiment_asr_mt} to generate 200-word summaries from ground truth conversations because the difference in summary style across prompts is not significant. 

\subsection{Results and Analysis}


\begin{figure*}[th]
  \centering
  \includegraphics[width=1\linewidth]{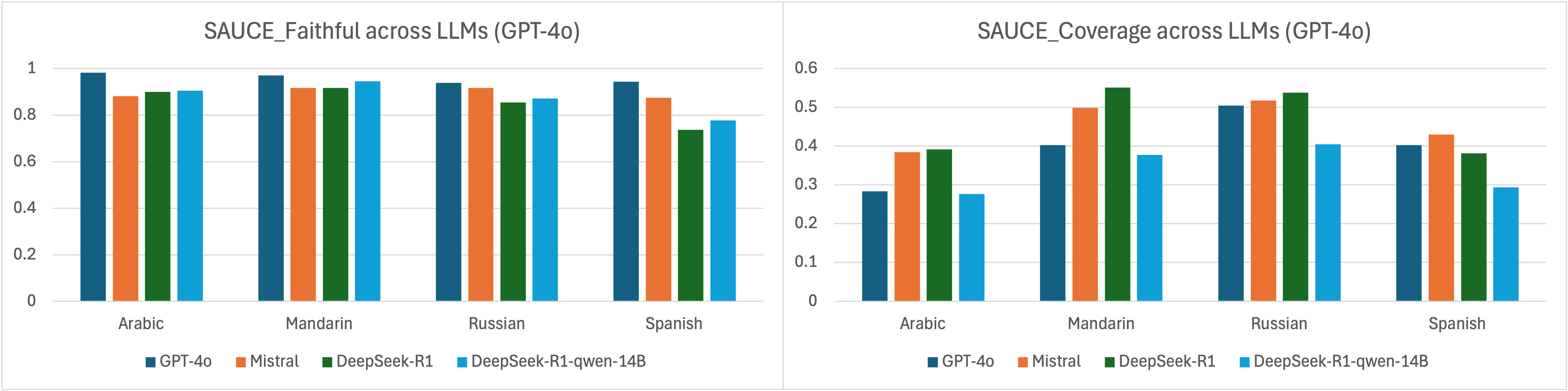}
  \caption{SAUCE\textsubscript{Faithful} and SAUCE\textsubscript{Coverage} on 4 LLMs across 4 languages translated into English.}
  \label{fig:LLM_scores}
\end{figure*}


Figure \ref{fig:LLM_scores} shows the average SAUCE\textsubscript{Faithful} and SAUCE\textsubscript{Coverage} of four off‑the‑shelf LLMs across English translated from four languages. GPT‑4o achieves the highest SAUCE\textsubscript{Faithful}, while DeepSeek attains the highest SAUCE\textsubscript{Coverage}. Coverage degrades after distilling DeepSeek‑R1 to a smaller 14B model. Surprisingly, DeepSeek‑R1’s SAUCE\textsubscript{Faithful} is the lowest, even lower than its distilled variant. SAUCE\textsubscript{Coverage} of GPT‐4o is less than that of a much smaller Mistral‑7B and is sometimes only comparable to DeepSeek‑14B in Arabic and Mandarin. These results suggest that DeepSeek‑R1 is less accurate than its distilled version, and although GPT‑4o produces the most accurate summaries, they cover less detail than those of other models.

We analyzed the summaries and identified several model behaviors that explain the observed scores (detailed examples and analysis are in Appendix Section \ref{sec:LLM_summary_detail}).
Misspellings and incorrect information lower faithfulness, while omitting subject details reduces coverage. We observed that GPT-4o summarizes the information both globally and locally; specific statements are compressed into more generic wording. This aligns with its high SAUCE\textsubscript{Faithful} but lower SAUCE\textsubscript{Coverage}. In contrast, Mistral7B and DeepSeek-R1 tend to preserve specific wording, producing extractive‑like summaries. However, hallucination occurs more often, introducing misinterpreted details or information that never appeared in the conversation. This reflects the higher SAUCE\textsubscript{Coverage} but lower SAUCE\textsubscript{Faithful} scores. Distilled DeepSeek contains less out-of-scope details its full model, also reflected in its higher SAUCE\textsubscript{Faithful}. Smaller model may lack the capacity for interpretation, making it more likely to quote directly from the source text.  


Ultimately, the explainability of SAUCE transforms model evaluation from a ranking exercise into a diagnostic one. Because every score is tied to discrete facts, we know exactly why DeepSeek's SAUCE\textsubscript{Faithful} is lower (it hallucinates non-existent conversational filler) and why GPT-4o's SAUCE\textsubscript{Coverage} is lower (it over-compresses specific events into abstract concepts). This fact-based traceability allows developers to make highly informed trade-offs based on their specific application needs, such as choosing Mistral-7B for high-coverage extractive tasks or GPT-4o for SAUCE, high-faithfulness abstractive overviews.
\subsection{Human evaluation on summaries quality}


\begin{table}[]  \centering
\begin{tabular}{c|cccc} 
\toprule
Coef.  & SAUCE\textsubscript{F}       & SAUCE\textsubscript{C}          & SC\textsubscript{Conv}   & SC\textsubscript{ZS}     \\\midrule
Acc. & \textbf{0.259} & 0.203          & 0.228 & 0.139          \\
Inf.    & 0.185          & \textbf{0.475} & 0.088          & 0.028          \\
\bottomrule
\end{tabular} 
\caption{Tau coefficient on human ranking of summaries compared with metrics. Notations as follow: \emph{Coef.}- Kendall rank coefficient, \emph{SAUCE\textsubscript{F}}- SAUCE\textsubscript{Faithful}, \emph{SAUCE\textsubscript{C}}- SAUCE\textsubscript{Coverage}, \emph{SC}- SummaC, \emph{Acc}- Accuracy, \emph{Inf.}- Informative.}
\label{tab:human_LLM}
\vskip -1em
\end{table}



We also performed an internal evaluation on the metrics against human intuition. We asked 12 participants to rank a small set of summaries generated by the 4 LLMs, based on accuracy and informativeness, resulting in 132 rank pairs. We calculate the Kendall's rank coefficient on the human ranked pairs and metrics ranked pairs in Table \ref{tab:human_LLM}. Pairs with less than a 0.05 score difference are treated as a tie. This is a difficult task and inter-conversation ranking pair agreement is only 60\% between participants. In Table \ref{tab:human_LLM}, SAUCE\textsubscript{Coverage} is highly correlated with informativeness and SAUCE\textsubscript{Faithful} is most correlated to accuracy, indicating our metrics align with human judgment. 

\section{Conclusion and Future Work} \label{sec:conclusion}

We propose a set of fact-based metrics for cross-lingual conversational speech summarization. We define SAUCE\textsubscript{Faithful} and SAUCE\textsubscript{Coverage} metrics to measure the accuracy and coverage of summaries by extracting facts and evaluating their entailment. Experiments demonstrate that our metrics outperform existing metrics in conversation summarization with the ability to trace back how the score are given, particularly in the presence of ASR and MT errors. We also evaluate multiple LLMs, showing that our metrics provide analytical insights into LLM performance. Future work includes improving the consistency of the fact extractor to mitigate potential fluctuation and bias, and developing more robust entailment models for conversational text.

\section{Limitations} \label{sec:limitation} 
The entailment model used in this study (DeBERTa-v3) was pre-trained on general text data and is not specifically tailored for conversational data, for which no dedicated entailment model currently exists. Future work involves fine-tuning DeBERTa-v3 for conversational entailment. However, this requires substantial amounts of annotated data that are presently unavailable.

Regarding fact extraction, our experiments relied solely on GPT-4o, which may be excessive for this task and carries the disadvantage of requiring API access, preventing offline deployment. Given that different LLMs exhibit distinct behaviors, future studies should evaluate various fact extractors and their potential biases. Our next step is to identify smaller LLMs for the pipeline and distill them using outputs from high-performing teacher models.

For our human evaluation of summary quality, the sample size was limited by resource constraints. Feedback from annotators indicated that reviewing long, transcribed conversations against summaries is exceptionally demanding. The inherent difficulty of systemic human cross-validation for such automatic pipelines poses challenges in precisely validating the overall performance.

\section{Ethics statement}

Due to the use of LLMs, ASR and MT, our metric can be subject to the bias of those models.
SAUCE provides a chain of artifacts so that a user can potentially discover any underlying bias in the final scores.
All audio recordings used in this study have been previously published.
Informed consent was obtained from all participants included in the study.

\bibliography{custom}

\appendix

\section{Extended Cross-Lingual Evaluation Metrics}
\label{sec:appendix_raw_results}

This section presents the comprehensive scoring breakdown for the evaluation frameworks discussed in this work. Table \ref{tab:raw_results_full} details the performance of SAUCE alongside established NLI-based and LLM-as-a-judge baselines—FineSurE, G-EVAL, FENICE, and SummaC—across Arabic, Mandarin, Russian, and Spanish. These results highlight the comparative stability of fact-based metrics when subject to the cascading noise of cascaded ASR with MT. While traditional metrics often struggle to generalize across diverse linguistic structures, the raw data support our observation that SAUCE provides a more consistent reflection of information loss and factual accuracy in conversational telephony speech.

\begin{table*}[h]
\centering
\small

\resizebox{\textwidth}{!}{
\begin{tabular}{llcccccccccccc}
\toprule
& & \multicolumn{3}{c}{Arabic} & \multicolumn{3}{c}{Mandarin} & \multicolumn{3}{c}{Russian} & \multicolumn{3}{c}{Spanish} \\
\cmidrule(lr){3-5} \cmidrule(lr){6-8} \cmidrule(lr){9-11} \cmidrule(lr){12-14}
Metric & Sub-metric & GT & MT & ASR & GT & MT & ASR & GT & MT & ASR & GT & MT & ASR \\
\midrule
\multirow{3}{*}{SummaC} & SummaC Conv & 0.237 & 0.230 & 0.226 & 0.212 & 0.213 & 0.212 & 0.217 & 0.217 & 0.213 & 0.758 & 0.769 & 0.789 \\
 & SummaC ZS & 0.385 & 0.363 & 0.364 & 0.298 & 0.305 & 0.280 & 0.318 & 0.312 & 0.327 & 0.325 & 0.313 & 0.326 \\
 & \textbf{AVG} & 0.311 & 0.296 & 0.295 & 0.255 & 0.259 & 0.246 & 0.268 & 0.264 & 0.270 & 0.541 & 0.541 & 0.558 \\
\midrule
\multirow{5}{*}{G-EVAL} & coherence & 0.810 & 0.800 & 0.755 & 0.890 & 0.870 & 0.857 & 0.929 & 0.900 & 0.883 & 0.920 & 0.894 & 0.870 \\
 & consistency & 0.875 & 0.735 & 0.620 & 0.907 & 0.747 & 0.660 & 0.925 & 0.825 & 0.771 & 0.902 & 0.812 & 0.736 \\
 & fluency & 0.450 & 0.515 & 0.500 & 0.493 & 0.463 & 0.347 & 0.475 & 0.433 & 0.367 & 0.494 & 0.454 & 0.460 \\
 & relevance & 0.765 & 0.755 & 0.655 & 0.813 & 0.777 & 0.720 & 0.821 & 0.812 & 0.754 & 0.824 & 0.796 & 0.770 \\
 & \textbf{AVG} & 0.725 & 0.701 & 0.633 & 0.776 & 0.714 & 0.646 & 0.787 & 0.743 & 0.694 & 0.785 & 0.739 & 0.709 \\
\midrule
\multirow{4}{*}{FineSurE} & faithfulness & 0.924 & 0.783 & 0.702 & 0.918 & 0.798 & 0.744 & 0.902 & 0.800 & 0.761 & 0.910 & 0.856 & 0.788 \\
 & completeness & 1.000 & 0.501 & 0.412 & 1.000 & 0.498 & 0.410 & 1.000 & 0.603 & 0.467 & 1.000 & 0.516 & 0.488 \\
 & conciseness & 0.946 & 0.758 & 0.644 & 0.881 & 0.628 & 0.541 & 0.949 & 0.815 & 0.704 & 0.934 & 0.720 & 0.691 \\
 & \textbf{AVG} & 0.957 & 0.681 & 0.586 & 0.933 & 0.641 & 0.565 & 0.950 & 0.739 & 0.644 & 0.948 & 0.697 & 0.656 \\
\midrule
\multirow{1}{*}{FENICE} & Score & 0.955 & 0.934 & 0.903 & 0.961 & 0.942 & 0.930 & 0.938 & 0.936 & 0.925 & 0.895 & 0.887 & 0.890 \\
\midrule
\multirow{3}{*}{\textbf{SAUCE (ours)}} & Faithful & 0.982 & 0.969 & 0.949 & 0.969 & 0.953 & 0.933 & 0.938 & 0.916 & 0.915 & 0.944 & 0.921 & 0.917 \\
 & Coverage & 0.284 & 0.221 & 0.169 & 0.402 & 0.364 & 0.350 & 0.504 & 0.427 & 0.368 & 0.403 & 0.337 & 0.284 \\
 & \textbf{AVG} & 0.633 & 0.595 & 0.559 & 0.686 & 0.658 & 0.641 & 0.721 & 0.671 & 0.641 & 0.673 & 0.629 & 0.601 \\
\bottomrule
\end{tabular}
}
\caption{Full Benchmark Results for Cross-Lingual Conversational Summarization Metrics Under Varied Noise Conditions. Scores evaluate summary quality across Ground Truth (GT), Machine Translation (MT), and cascaded ASR-MT inputs, highlighting the robustness of SAUCE (Faithfulness and Coverage) compared to existing baselines.}
\label{tab:raw_results_full}
\end{table*}

\section{Detail analysis on summarizing ASR+MT conversations}
\subsection{ASR and MT performance}
\label{sec:asr_mt_detail}

Table \ref{tab:asr_mt} provides a detailed breakdown on the WER and BLEU scores in Figure \ref{fig:asr_mt_summ_score}. It reflects the degradation of transcription quality resulting from the use of ASR and MT on cross-lingual conversational data.
\begin{table*}[ht] 
\centering
\begin{tabular}{cccc}
\hline
 \textbf{Language} & \textbf{ASR (WER)} & \textbf{MT (BLEU)} & \textbf{ASR+MT(BLEU)} \\ \hline
\textbf{Arabic (avg)}        & \textbf{0.497}     & \textbf{0.314}     & \textbf{0.211}        \\ \hline
CallHome dev           & 0.507     & 0.311     & 0.233        \\
CallHome test          & 0.487     & 0.317     & 0.189        \\ \hline
\textbf{Mandarin (avg)}   & \textbf{0.341}     & \textbf{0.249}     & \textbf{0.177}        \\ \hline
HUB5 dev           & 0.349     & 0.261     & 0.178        \\
HUB5 test          & 0.333     & 0.237     & 0.177        \\ \hline
\textbf{Russian (avg)}   & \textbf{0.343}     & \textbf{0.341}     & \textbf{0.211}        \\ \hline
Mixer3 dev           & 0.351     & 0.345     & 0.214        \\
Mixer3 test          & 0.334     & 0.338     & 0.209        \\ \hline
\textbf{Spanish  (avg)}   & \textbf{0.229}     & \textbf{0.313}     & \textbf{0.240}        \\ \hline
CallHome dev  & 0.218     & 0.303     & 0.226        \\
CallHome test & 0.230     & 0.312     & 0.230        \\
Fisher dev    & 0.226     & 0.313     & 0.226        \\
Fisher dev2   & 0.236     & 0.334     & 0.236        \\
Fisher test   & 0.234     & 0.314     & 0.264       \\ \hline
\end{tabular} 
\caption{WER and BLEU scores for each language on average, and their corresponding dev and test sets.}
\label{tab:asr_mt}
\end{table*}


\subsection{Conversation facts extracted from sp\_1578}
\label{sec:fact_examples}

Table \ref{tab:spanish_convo_fact_example} presents a subset of facts extracted from the ground truth translated conversation from sp\_1578. The facts are highlighted and labeled according to specific sets of information in Table \ref{tab:spanish_convo_example}. Noticed the variations on the conversation facts regarding the same information, therefore we extract and concatenate multiple sets of facts for SAUCE\textsubscript{Coverage} scoring for a stable performance leading.

\begin{table*}[]
\begin{tabular}{p{0.95\linewidth} } 
\toprule
\textbf{Subset of facts extracted from the conversation} \\

\midrule
{\color{blue} Facts (a): Overview of conversation.} \\
1. Carlos cannot undergo surgery until his health improves. \\
2. Someone had a total body allergy from pills Cristi bought. \\
3. Someone is allergic to antibiotics and cannot self-medicate. \\
4. A family is fixing their house and making it bigger. \\
5. The family is going to put a store in the house. \\ \hline

{\color{purple} Facts (b): Carlos' interaction with speaker.} \\
1. Carlos was not given anything during the visit. \\
2. Carlos drove the person speaking to Joana's place. \\
3. Carlos drove A. \\
4. The speaker cannot leave Carlos because he is not having a good time. \\ 
5. Carlos needs more support, especially moral support. \\ 
6. Carlos is not having a good time and needs support. \\
\hline
{\color{orange} Facts (c): Ernesto driving.} \\
1. Ernesto bought a car.\\
2. Ernesto bought a "super cute little car." \\
3. Ernesto is learning to drive.\\
4. Ernesto is going to take a driving course.\\ \hline
{\color{olive} Facts (d): Carlos' health.} \\
1. Carlos has symptoms similar to sinusitis, with nose issues and sneezing. \\
2. Carlos has an allergy similar to sinusitis. \\
3. Carlos' operation is delayed until he is well. \\
4. B went to the doctor due to a cold. \\
5. Carlos's surgery was postponed to November due to a cold related to an allergy. \\
6. The surgery was postponed because Carlos has an allergy and a cold.\\ 
7. B is allergic to antibiotics. \\
8. Someone was scolded by a doctor for taking unprescribed medication. \\ \hline
\end{tabular} 
\caption{Example of Facts extracted Spanish conversation (sp\_1578) on ground truth translation, grouped by the relevant topics. The colors and alphabets scheme aligns with Table \ref{tab:spanish_convo_example}.}
\label{tab:spanish_convo_fact_example}
\vskip -0.5em
\end{table*}

\section{Detail analysis on summaries generated by various LLMs}
\label{sec:LLM_summary_detail}

We analyzed the summaries and presented some examples of SAUCE\textsubscript{Coverage} facts and summary sentences in Table \ref{tab:LLM_summary}. We highlighted correct information in blue, incorrect in red, and over-interpretation in orange. 
As discussed in earlier section, GPT-4o generate abstractive summary with higher accuracy, while Mistral generates extractive summary with specifics, with more error. For Fact 1, Mistral7B got the incorrect name, but describes all the sports activities involved. GPT-4o got the name correct, but generalized all activities to ``sports". DeepSeek models mentioned part of the sport activities, but misinterpreted and included extra information that is not in the conversation.
For Fact 2, GPT-4o generalized it to ``securing a green card" without giving much detail. Other models correctly describe the stage of the green card process, but Mistral7B does not name what process it is. DeepSeek provides a reason for approval that is not found in conversation. Misspelling and wrong information leads to lower SAUCE\textsubscript{Faithful} and missing the subject detail leads to lower SAUCE\textsubscript{Coverage}, this explaining the lower SAUCE\textsubscript{Faithful} for Mistral7B and DeepSeek, and lower SAUCE\textsubscript{Coverage} for GPT-4o. This explains the scores on the models. 

\begin{table*}[ht] 

\begin{tabular}{p{0.95\linewidth} } \hline

\textbf{Arabic conversation (ar\_4264}) \\ \hline
\textbf{Fact 1.1:} Ashraf is learning baseball, TUI, karate, and basketball at the summer school. \\
\textbf{Fact 1.2:} A decided to stop swimming lessons for the summer. \\
\textbf{Related conversation line:} A: well I stopped it in Summer because I go to swim with these Muslim sisters
\\ \hline

\textbf{GPT:} A shares updates on {\color{blue}Ashraf's} activities in a summer program learning {\color{blue}various sports}. \\ 
\textbf{Mistral:} A also mentioned that their son, {\color{red}Ahmad}, is doing well and attending a summer school program where he is learning {\color{blue}baseball, TUI, karate, basketball, and possibly karate.} \\
\textbf{DeepSeek:} Updates on family members follow, including {\color{blue}Ashraf} attending summer school for activities like {\color{blue}baseball and karate} before the regular school term starts, {\color{red} while swimming lessons are paused}. \\
\textbf{DeepSeek-distil:} The discussion shifts to the children, particularly {\color{blue}Ashraf}, who is attending summer school to learn sports like {\color{blue}baseball, karate, and basketball}. A mentions stopping swimming lessons {\color{red}due to religious reasons}. \\
\hline
\textbf{Fact 2:}  A green card process is underway; the first step approved. \\ \hline
\textbf{GPT:} There's a detailed discussion about securing a {\color{blue}green card}, the importance of consistent work and tax records, and academic goals \\ 
\textbf{Mistral:} Regarding a family matter, A mentioned that they are waiting for a response from the committee regarding a {\color{blue}second step in a process}, which is crucial for them. \\
\textbf{DeepSeek:} A significant portion focuses on the complex {\color{blue}green card} application process; the initial step is approved {\color{red}based on the husband's unique skills}, but {\color{orange}anxiety} surrounds the crucial {\color{blue}second step} involving lifetime background checks, tax history, and the potential financial burden. \\
\textbf{DeepSeek-distil:} A turns to the challenges of the {\color{blue}green card} process, particularly the {\color{blue}second step} for work authorization, which is crucial but {\color{orange}frustrating}. \\
\hline
 \textbf{Mandarin conversation (ma\_4249)} \\ \hline
\textbf{Fact 3.1:} B received a letter congratulating her daughter on being selected as a model. \\
\textbf{Fact 3.2:} B's daughter, Helen, was selected as a baby model. \\
\hline
\textbf{GPT:} There is a mention of a {\color{blue}child being selected as a model} and the evaluations that follow. \\ 
\textbf{Mistral:} B mentions that their daughter, {\color{red}Helena, is a model baby} and has been {\color{blue}selected as a model}. \\
\textbf{DeepSeek:} and mentions {\color{blue}Helen’s recent selection as a baby model} {\color{red}after a radio contest}...The conversation underscores the exhaustion of balancing childcare with personal goals, alongside {\color{orange}fleeting moments of pride in their children’s milestones.}  \\
\textbf{DeepSeek-distil:} The conversation concludes with excitement over Helen's {\color{blue}selection as a model}, with {\color{blue}B receiving a letter} about the opportunity, {\color{orange}though details remain unclear}. \\
\hline
\end{tabular} 
\caption{Chunks of summaries with their corresponding facts on one Arabic conversation (ar\_4264) and Mandarin conversation (ma\_4249) on GPT, Mistral7B, DeepSeek-R1 and DeepSeek-R1-distil-qwen-14B. Blue indicates the information that are aligned with facts, red indicates incorrect information, orange indicates non-factual interpretation.}
\label{tab:LLM_summary}
\end{table*}


\end{document}